\documentclass[runningheads,a4paper]{llncs}

\usepackage{amssymb}
\usepackage{graphicx}
\usepackage{amsmath}
\usepackage{amsfonts}

\usepackage{url}
\urldef{\mailsa}\path|{antonio.zippo@gmail.com, christine.guenther,|
\urldef{\mailsb}\path|ingrid.haas, frank.holzwarth, anna.kramer, leonie.kunz, nicole.sator,|
\urldef{\mailsc}\path|erika.siebert-cole, peter.strasser, lncs}@springer.com|
\newcommand{\keywords}[1]{\par\addvspace\baselineskip
\noindent\keywordname\enspace\ignorespaces#1}

\begin{document}

\mainmatter  
\title{Handwritten digits recognition by bio-inspired hierarchical networks}

\author{Antonio G. Zippo \and Giuliana Gelsomino \and Sara Nencini \and Gabriele E. M. Biella}

\authorrunning{Handwritten digit recognition by bio-inspired hierarchical networks}

\institute{Institute of Molecular Bioimaging and Physiology,\\
Consiglio Nazionale delle Ricerche,\\
Via Fratelli Cervi 21, 20090 Segrate (Milan), Italy\\
\url{http://www.ibfm.cnr.it}}

\toctitle{Lecture Notes in Computer Science}
\tocauthor{Authors' Instructions}
\maketitle

\begin{abstract}
The human brain processes information showing learning and prediction abilities but the underlying neuronal mechanisms still remain unknown. Recently, many studies prove that neuronal networks are able of both generalizations and associations of sensory inputs.\\ 
In this paper, following a set of neurophysiological evidences, we propose a learning framework with a strong biological plausibility that mimics prominent functions of cortical circuitries. We developed the Inductive Conceptual Network (ICN), that is a hierarchical bio-inspired network, able to learn invariant patterns by Variable-order Markov Models implemented in its nodes. The outputs of the top-most node of ICN hierarchy, representing the highest input generalization, allow for automatic classification of inputs. We found that the ICN clusterized MNIST images with an error of 5.73\% and USPS images with an error of 12.56\%.  

\end{abstract}

\section{Introduction}
The brain is a computational device for information processing and its flexible and adaptive behaviors emerge from a system of interacting neurons depicting very complex networks \cite{bassett}. Many biological evidences suggest that the neocortex implements a common set of algorithms to perform ``intelligent'' behaviors like learning and prediction. In particular, two important related aspects seem to represent the crucial core for learning in biological neural networks: the hierarchical information processing and the abstraction process \cite{dicarlo}. The hierarchical architecture emerges from anatomical considerations and is fundamental for associative learning (e.g. multisensory integration). The abstraction instead leads the inference of concepts from senses and perceptions (Fig. \ref{fig:1}D).\\
Specifically, information from sensory receptors (eyes, skin, ears, etc.) travels into the human cortical circuits following subsequent abstraction processes. For instance, elementary sound features (e.g. frequency, intensity, etc.) are first processed in the primary stages of human auditory system (choclea). Subsequently sound information gets all the stages of the auditory pathway up to the cortex where higher level features are extracted (Fig. \ref{fig:1}E-F). In this way information passes from raw data to objects, following an abstraction process in a hierarchical layout. Thus, biological neural networks perform generalization and association of sensory information. For instance, we can associate sounds, images or other sensory objects that present together as it happens in many natural and experimental settings like during Pavlovian conditioning. Biological networks process these inputs following a hierarchical order. In a first stations inputs from distinct senses are separately processed accomplishing data abstraction. This process is repeated in each subsequent higher hierarchical layer. Doing so, in some hierarchical layer, inputs from several senses converge showing associations among sensory inputs.\\
Recent findings indicate that neurons can perform invariant recognitions of their input activity patterns producing specific modulations of their synaptic releases \cite{takahashi,kleindienst,makino,meyers,branco}. Although the comphrension of such neuronal mechanisms is still elusive, these hints can drive the development of algorithms closer to biology than spiking networks or other brain-inspired models appear to be.\\ 
In this work, we propose a learning framework based on these biological considerations, called Inductive Conceptual Network (ICN), and we tested the accuracy of this network on the MNIST and USPS datasets. The ICN represents a general biological plausible model of the learning mechanisms in neuronal networks. The invariant pattern recognition that occurs in the hierarchy nodes is achieved by modeling node inputs by Variable-order Markov Models (VMMs) \cite{buhlmann,begleiter}.

\section{Methods}\label{Methods}
The methods of this work are based on a set of considerations extracted primarily from the Memory-Prediction framework proposed by Jeff Hawkins in his book \textit{On Intelligence}. Therefore in this section we first present crucial aspects of brain information processing.

\subsection{Background about learning and the Memory-Prediction Framework}
As preliminary step we introduce few theoretical concepts about learning and memory experiences in nervous systems. The human brain massively elaborates sensory information. Through some elusive mechanism, the brain builds models (formal representation) from observations. In such models, pattern recognition and abstraction play a crucial role \cite{hawkins}. The former allows for the capture of patterns from observations, the latter allows for transforming raw observations into abstract concepts. For instance, listening to sequence of unknown songs from an unknown singer we perform both pattern recognition and abstraction, respectively when we identify sound features (e.g. beats per minute) and when we infer abstract information concerning the new singer (e.g. he/she plays jazz).\\
Key features of these brain processes can be translated in algorithms \cite{dileep}. Jeff Hawkins et al. recently proposed a new learning framework (Memory-Prediction \cite{hawkins}) based on abstraction processes and pattern recognitions. This paradigm claims that abstraction represents one of the most important tasks underlying learning in the brain and that occurs through the recognition of invariances. Moreover, he suggested that sensory inputs are processed hierarchically: each layer propagates to the next layer the invariant recognized patterns. In propagating only invariances and discarding everything else, data are compressed with size decreasing at every next layer. This is finely promoted by a pyramidal shape. Hawkins et al. implemented the Memory-Prediction framework into a set of software libraries specialized in image processing (Hierarchical Temporal Memory, HTM \cite{dileep}) which exhibits invariant recognition by a complex hierarchy of node implementing the Hidden Markov Model algorithm \cite{baum1966,baum1970}.

\begin{figure}[ht]
\centering
\includegraphics[width=\textwidth]{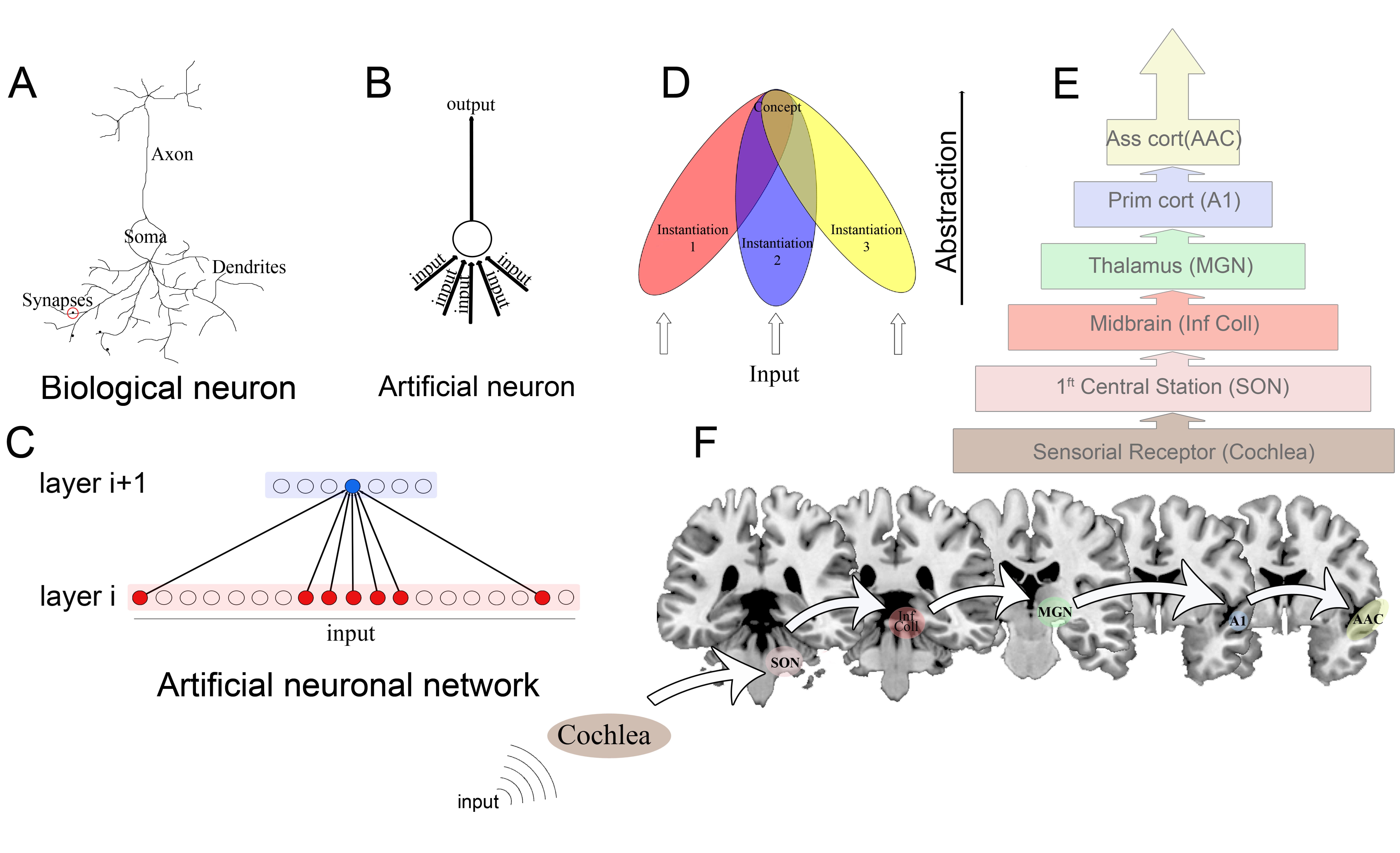}
\caption{Background and preliminary concepts on Inductive Conceptual Network (ICN). (A)-(B) Comparison between biological and artificial neurons. Biological signals conducted by each dendrite on soma can be represented by artificial inputs; after the input elaboration, axon conducts the signal output that in artificial neuron is computed by estimation of probability distribution of the observed inputs. (C) Graphically, neurons are represented by nodes (circle) which are organized in layers. They are linked by inter layer connections (edges) following a proximity criterion admitting exceptions. (D) Representation of the hierarchical abstraction framework that occurs getting from input representing raw data (concept instances, e.g. “one”, “seven”) to concepts (“number”). (E) An example of biological correspondence between the ICN and the auditory sensorial system in human. Auditory input elaborated from cochlea, through sensory pathway reaches auditory associative cortex. (F) The auditory sensory pathways seen in the coronal MRI template slices. Abbreviations: SON, Superior Olivary Complex; Inf Coll, Inferior Colliculi; MGN, Medial Geniculate Nucleus; A1, Primary auditory cortex, AAC, Auditory Associative Cortex.}\label{fig:1}
\end{figure}

\subsection{The Inductive Conceptual Network}
We propose a different realization of the Memory-Prediction framework, called Inductive Conceptual Network (ICN), where biological neurons are individually identified by nodes (see Figure \ref{fig:1}A-B) and invariant recognition is performed by modeling inputs with Variable-order Markov Models (VMM) \cite{buhlmann,begleiter,rissanen}. The former assumption allowed us to pin down the ICN model into adequate biological background and to evaluate not only its learning ability but also its neurophysiological matching with neuronal population dynamics. The latter assumption addresses the problem of invariant recognition in a powerful and computational efficient way \cite{begleiter}.\\ 
The Inductive Conceptual Network is a hierarchical spiking network working as online unsupervised learning algorithm. In common with HTM, the ICN displays a tree-shaped hierarchy of nodes (Figure \ref{fig:1}D-F). Formally, ICN is a triplet $(T,M,k)$ where $T = \{l_1,l_2,\dots,l_L\}$ is the vector that contains the number of nodes in each layer such that $l_1 > l_2 > \cdots > l_L = 1$. Let $q=\sum_{i=1}^L l_i$ be the total number of nodes, and $M$ is the $q$x$q$ adjacency matrix representing the connections between nodes and $k$ is the maximum Markov order, an indicator of the memory power of each node. For the construction of $M=\{m_{i,j}|i,j=1,\dots,q\}$ that is initially set to $m_{i,j}=0$, we proceeded iteratively following these two steps for nodes in each layer $x$:
\begin{enumerate}
\item a set of deterministic assignations: \{$m_{i,i+l_x}=1,\dots,m_{i+p,i+l_x}=1$\} with $p = \lceil\frac{l_x}{l_{x+1}}\rceil +1$ and $\forall i \in \{\sum_{k=1}^x l_x,\dots,(\sum_{k=1}^{x+1} l_x) - \lceil\frac{l_x}{l_{x+1}}\rceil \}$;
\item a set of random assignations: $\{m_{i+r,i+l_x}=1| r \sim U(1,l_{x_1})\}$ 
\end{enumerate}
\noindent where $l_x$ is the number of nodes in the generic layer $x$ and $U$ is the discrete uniform distribution. Layers handle inputs from the immediately preceding layer (layer below) except for the first that handles the raw input data. The matrix $M$ is semi-randomly assigned respecting the multilayer architecture: each node receives the downstairs-layer input both from their neighbour nodes and from a small set of randomly chosen ones (Figure \ref{fig:1}C).\\
Nodes read inputs from their \textit{dendrites} (Figs. \ref{fig:1}A-B) and an algorithm estimates the joint probability distribution of the observed inputs (see below, VMM). Whether the observed input is the most expected (or is very close to) the node produces a $1$ (representing a spike) towards their output nodes otherwise does nothing. The ICN is a general purpose learning framework, and although it has not been tested on non-visual tasks it can however be used for other sensory information processing.

\subsection{Variable-order Markov Models}
The learning of spatiotemporal patterns is the subject of study of Sequential data learning that usually involves very common methods, like Hidden Markov Models (HMM). In fact, HMM are able to model complex symbolic sequences assuming hidden states that control the system dynamics. However, HMM training suffers from local optima and their accuracy performance has been overcome by VMMs. Other techniques like $N$-gram models (or $N$-order Markov Models) compute the frequency of each $N$ long subsequence but in these models the number of possible model states grows exponentially with $N$. Therefore, both computational space and time issues arise.\\
In this perspective, the observed symbolic (binary) sequence is assumed to be generated by a stationary unknown symbol source $S=\left< \Sigma,P \right>$ where $\Sigma$ is the symbol alphabet and $P$ is the probability distribution of symbols. A VMM (also known as Variable length Markov Chains), given the maximum order $D$ of conditional dependencies and a training sequence $s$ generated by $S$, returns a model for the source $S$ that's an estimation $\hat{P}$ of probability distribution $P$. Applying VMMs, instead of $N$-gram models, takes several advantages. A VMM estimation algorithm builds efficiently a model for $S$. In fact, only the occurred $D$-grams are stored and their conditional probabilities $p(\sigma|s)\;,\sigma \in \Sigma\; \text{and } s\in \Sigma^{d \leq D}$ are estimated. This trick saves lots of memory and computational time and makes feasible to model sequences with very long dependencies ($D \in [1,10^3]$) on current personal computers.

\subsection{The node behavior and invariance recognition}\label{node_physiology}
We consider the inputs from dendrites that each neuron (node) sees as binary symbols emitted by a discrete source which releases outcomes following an unknown non-stationary probability distribution $P$. The aim of each node is to learn its source as best as possible so that it can recognize correctly recurrent patterns assigning to them highest probabilities. The VMMs are typically used for this task being able to model dependencies among symbols up to an arbitrary order. VMMs can be estimated by many algorithms. We took into consideration a famous efficient lossless compression algorithm, the Prediction by Partial Matching (PPM) \cite{cleary,teahan}, implemented in an open-source Java software library \cite{vmmrepository}.\\
Formally, a node reads a binary input (at each step) $s = (s_1,\dots,s_n)$ of length $n$ that represents the all-or-none dendritic activity. Let $k < n$ be the maximum dependency allowed among input symbols, then each node builds its probability model feeding $k$-tuples of the received $n$-ary input $s$ into the PPM algorithm. Each node has its own instance of the PPM algorithm. After this first learning phase, the node passes into the prediction mode and looks if it observes in $s$ the most expected pattern (pattern that has the highest probability assignment). If it happens, the node produces $1$ as output in correspondence to the salient patterns thus preserving the spatial structural organization of inputs. We introduce the further condition that a $1$ is produced in correspondence of patterns having Hamming distance \cite{hamming} very close to the most expected one. We make this choice to introduce a sort of noise tolerance in the pattern recognition process. In other words, during the coding (and second) stage, a node processes its input by $k$ symbols at time. If the current $k$-tuple pattern is the highest probable (or is very close to, by Hamming distance) a $1$ is inserted into the output code, otherwise it marks a $0$.\\
For instance, let be $k=3$ and $101$ the most expected pattern. Let $11000010\-1\-1\-0\-0\-1\-01$ be the current input that updates the probability distribution $P$. Finally, the node produces the output sequence $00101$ where $1$ corresponds to the two occurrences of the most expected pattern ($101$). The pseudo-code of the algorithms governing respectively nodes and the hierarchy are the following:

\medskip

\noindent
{\it Algorithm for nodes}
\begin{verbatim}
Node()
   read input s = (s_1,...,s_n);
   for each k-tuple in s:
      update P by PPM(s_i,...,s_(i+k));
      if HammingDistance((s_i,...,s_(i+k)),bestPattern) < gamma:
            output(1);
      else
            output(0);
      update bestPattern;
   end
end
\end{verbatim}
where the function \texttt{PPM()} updates the probabilistic model $\hat{P}$ with the new input s\_i,...,s\_(i+k). The function \texttt{HammingDistance($\cdot$,$\cdot$)} computes the Hamming distance between two binary strings and the function \texttt{PPM\_best()} returns the current most probable pattern.

\medskip

\begin{figure}[ht!]
\centering
\includegraphics[width=\textwidth]{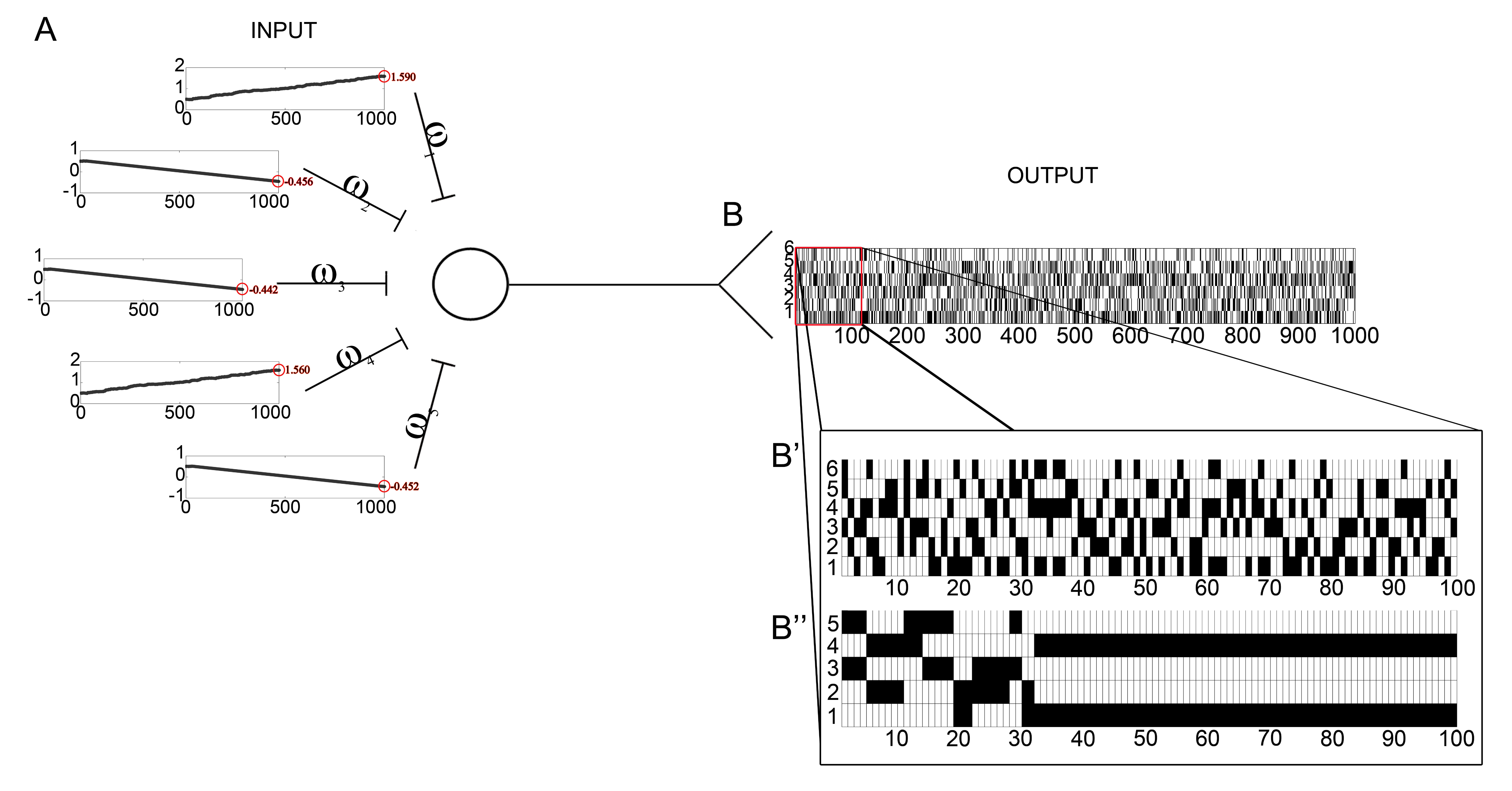}
\caption{Learning in a node of ICN assuming synaptic weights ($\omega$) and the Hebb$^\prime$s rule. (A) Starting with equal weights (0.5) and assigning a positive increment (0.01) whether pre-synaptic spike precedes the post-synaptic spike in 2 timesteps at most. Otherwise the synaptic weight incurs in a negative reward (-0.01). The sequence of input patterns is composed by randomly generated binary inputs (with probability 0.75) plus a fixed input equal to $10010$ (with probability 0.25) . The simulation lasts 1000 timesteps where, at the end, the recurrent pattern $10010$ was recognized assigning strong weights to the first and fourth synapses depressing the other ones. (B) In detail, the raster plot of the simulation where the activity of nodes 1-5 matches the activity of the 5 presynaptic inputs and the activity of node 6 is the output of node in examination. (B$^\prime$) An enlargement of the first 100 timesteps. (B$^{\prime\prime}$) The evolution of the most expected pattern according to the PPM estimation in the node. After the first 31 timesteps, the fixed pattern $10010$ becomes the most expected.}\label{fig:2}
\end{figure}

\noindent
{\it Algorithm for ICN}
\begin{verbatim}
ICN()
  for each image in dataset:
      bw = binarizeImage(image);
      assignInputToFirstLayer(bw);
      foreach layer in ICN:
           setInput(bw);
           learn();
           bw = predict();
      end
      collect(bw)
  end
end
\end{verbatim}
\noindent where in the \texttt{learn()} function the distribution $\hat{P}$ of each node are updated and in the \texttt{predict()} function the spiking activity of the current layer is returned.\\
Before evaluating the performance of ICN on handwritten digit datasets, we evaluated the learning capabilities of a single node by a simple experiment. We provided a sequence of 1000 binary 5-tuples as input to a node with 5 dendrites and $k=5$ (Figs. 2A-B). The input sequence of 5-tuple is generated randomly inserting at each time with probability 0.25 a fixed pattern (equal to $10010$). Simulating a Hebbian setup where at each of the five dendrites is associated a weight increased in case of positive temporal correlation between pre- and post-synaptic spikes and decreased otherwise \cite{hebb}, we make sure that the end of proposed sequence weights of first and fourth dendrites are strengthen to the detriment of other ones (Figure \ref{fig:2}A).

\subsection{Learning of handwritten digits}
In the current form, ICN can perform unsupervised learning. To evaluate the learning capabilities of such framework, we gave as input to the first layer, the images of the MNIST (or USPS) dataset (handwritten digits, 0 to 9). Here we use the MNIST test set which contains 10000 examples and the whole 11000 sample of the USPS. The chosen instantiation of ICN was composed by 4 layers with respectively 50, 20, 5 and 1 node in each layer. The maximum Markov order $k$ was set to $5$ for all ICN nodes. All parameters in this section have been chosen empirically to best match the right classification of the digits. We expected that digit images were correctly grouped with respect to the represented number. MNIST images are represented by 28x28 (784) pixels of 8-bit gray level matrix. Instead, USPS images, are represented by 16x16 (256) pixels. Images were binarized setting a threshold on the 8-bit gray-level values to 80. As explained above, nodes produce bits and the result of this unsupervised learning is valuable in the outputs of the top-most node. In fact, this node retains the most abstract information regarding the observed images. Namely, something likes the concept of number. After some empirical tuning of parameters (number of nodes, layer and maximum Markov order), ICN was able to discriminate digits by the top-most node output code. For instance in some experiments, giving an image of digit 0, the ICN emitted the binary code $1000$. In the same experiments, the code $0101$ was reserved to the digit 1 and so on. Obviously, the ICN made errors and digit-to-code associations were not unique, e.g. some seven digits can be incorrectly classified with code $0101$ reserved for the 1. To estimate the learning error, we chose a representative code for each digit class. The representatives were selected as most frequents for each class. Thus, the learning error was computed by counting mismatchs between labels and representative codes.

\section{Experimental results}\label{experiments}
The ICN algorithm has been developed following strict and recently found biological criteria from the neurophysiology of neuronal networks. Once ascertained that ICN nodes perform a sort of Hebbian plasticity (see section \ref{node_physiology}) we challenged the ICN with the MNIST dataset (handwritten digit images). The MNIST dataset represents a sort of \textit{casting-out-nines} for learning systems; in fact, new proposed algorithms are tested on this dataset to check their attitude to learn.\\
The learning capabilities of ICN were tested by its clustering efficiency over the MNIST dataset. Before submitted to ICN every digit image was binarized by applying a threshold. Subsequently each image was fed into the first layer nodes. Invariant recognized patterns are then propagated, layer-by-layer up to the highest, following the execution of Algorithm-2 (see Methods). As a whole, an input image elicits a bit (spike) flux in the bottom layer, a code transmitted to the upper layer. The top-most layer, composed by only one node, finally generates its binary codes each corresponding to a digit (class) of the image input. 
We ascertained that at the best tuning of parameters the ICN model got an average error of 5.73\%, an acceptable score in an unsupervised environment, remarkably not requiring any preprocessing stages such as image alignment, centering or dimensionality reduction. For the USPS dataset, however harder to learn, the best achieved error was of 12.56\%.\\
Eventually, we further investigated the influence of dataset size in the learning performance. For this reason, we repeated the same experiments randomly subsampling both datasets to 1000 and 5000 samples. For both datasets, performance improved increasing the dataset size as shown in table \ref{tb1}.

\begin{figure}[ht!]
\centering
\includegraphics[scale=0.045]{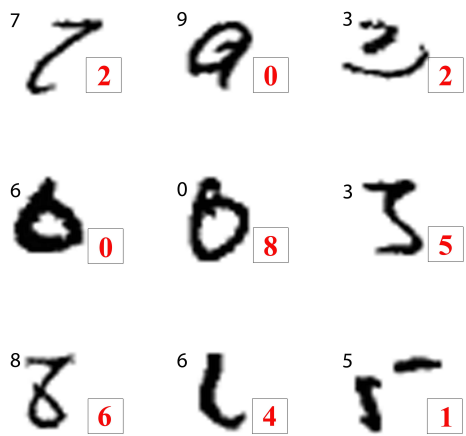}
\caption{Sample of common incorrect classifications on MNIST dataset. Numbers in the upper left of boxes indicate the correct representation. Numbers in the lower right of boxes indicate the incorrect classifications.}\label{fig:3}
\end{figure}


\begin{figure}[ht!]
\centering
\includegraphics[width=\textwidth]{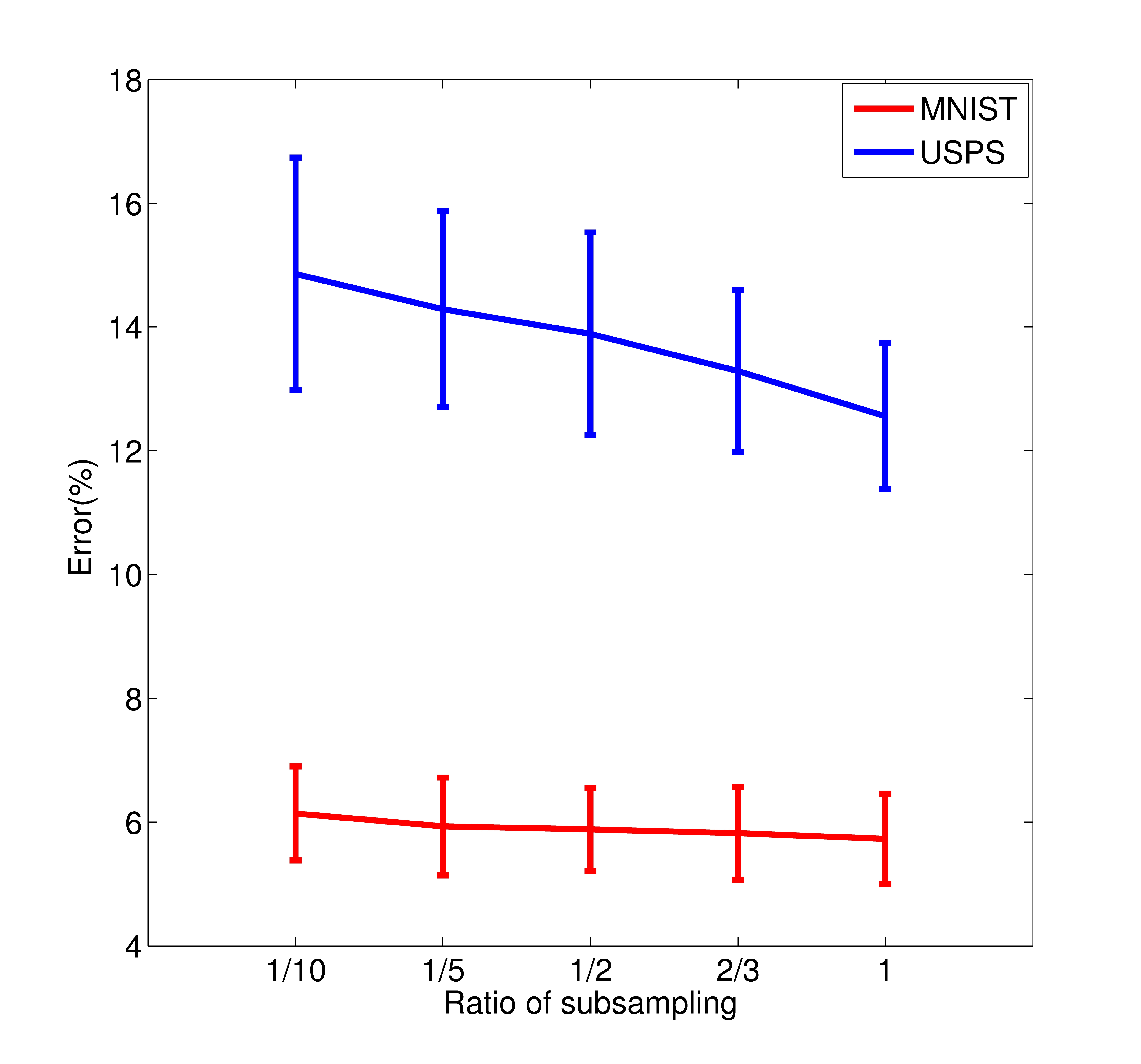}
\caption{Learning performances computed on 100 trials for each subsampling of the original datasets.}\label{tb1}
\end{figure}

\section{Discussion}
Even convolutional neural networks (CNNs) \cite{lecun1990,lecun1995} are biologically inspired by the pioneer works of Hubel et al. on the retinotopies of the cat's visual cortex \cite{hubel}. Indeed, CNNs exploit the fact that nearby pixels are more tightly correlated than more distant ones. Furthermore by using a cascading structure of convolutional and subsampling layers, these networks show successfully invariant recognition of handwritten digits subjected to certain transformations (scaling, rotation or elastic deformation). Altough CNNs are bio-inspired by the local receptive fields which constitute the local features, the learning mechanism of the whole network does not appear to have a biological counterpart. Vice versa, the proposed network (ICN) implements invariant recognition exhibiting a spiking behavior in each node which represents a clear correspondence with biological networks. Furthermore, the algorithm governing nodes is the same in the whole network. Since the electrophysiological properties of neurons are quite similar, our network appears to be more plausible than CNNs where a set of special layers (and nodes) exclusively perform the invariant recognition.\\
The performance of each node is based on the PPM algorithm that requires $O(n)$ during learning and $O(n^2)$ during prediction as computational time complexities \cite{begleiter}. Although the quadratic complexity, each node receives only small fractions of inputs keeping $n$ within small values. Thus the overall time complexity for each processed image raises to $O(m \cdot n^2)$, where $m$ is the number of nodes. Interestingly, the node executions within each layer can be computed in parallel. Even the space complexity is dictated by the complexity of the PPM algorithm that is $O(k\cdot n)$, where $k$ is the chosen Markov order, in the worst case. Therefore, the ICN algorithm requires $O(m\cdot k\cdot n)$ in space complexity.

\section{Conclusion}\label{conclusion}
The MNIST dataset is a standard test to evaluate learning accuracy for both linear and non-linear classifiers. We show here that ICN is apt to carry out unsupervised learning tasks with an error rate of 5.73\% for MNIST and 12.56\% for USPS at most. The percentage may appear weaker, in comparison with other learning methods, seemingly showing better error rates thanks, however to training and preprocessing (check for instance the performance of convolutional nets scoring down to 0.35\% error rate). Furthermore, in comparison with other clustering techniques, our method does not fail into the \textit{curse of dimensionality} \cite{powell}. Any classical unsupervised learning techniques, such as k-means, Expectation-Maximization or Support Vector Machines generally require an \textit{ad hoc} dimensionality reduction (e.g. by Independent or Principal Component Analysis), a procedure that reduces the algorithm general purposiveness \cite{kotsiantis}. However, these networks do not acknowledge biological modeling, where ICN is instead adequately biologically oriented.\\
In conclusion, the proposed model achieves interesting preliminary results. Nevertheless further experiments with other machine learning datasets are required to strengthen its validity. Moreover, future developments can allow for effective multi-input integrations: for instance, two different sources of input (like sounds and images) could be associated by similar output codes even in presence of inputs from a single source.

\end{document}